# New Ideas for Brain Modelling 6

Kieran Greer, Distributed Computing Systems, Belfast, UK.

http://distributedcomputingsystems.co.uk

Version 1.0

*Abstract*—This paper describes implementation details for a 3-level cognitive model, described in the paper series. The whole architecture is now modular, with different levels using different types of information. The ensemble-hierarchy relationship is maintained and placed in the bottom optimising and middle aggregating levels, to store memory objects and their relations. The top-level cognitive layer has been re-designed to model the Cognitive Process Language (CPL) of an earlier paper, by refactoring it into a network structure with a light scheduler. The cortex brain region is thought to be hierarchical - clustering from simple to more complex features. The refactored network might therefore challenge conventional thinking on that brain region. It is also argued that the function and structure in particular, of the new top level, is similar to the psychology theory of chunking. The model is still only a framework and does not have enough information for real intelligence. But a framework is now implemented over the whole design and so can give a more complete picture about the potential for results.

## 1   Introduction

This paper describes a framework that has been implemented, using the bio-inspired design of the papers 'New Ideas in Brain Modelling 1-6' [6]-[10] and the original 3-level cognitive architecture [14][12]. The design is imagined to be as close to the human brain as possible, where there are different aspects of brain construction that can be modelled. This paper is interested in a general framework that can be filled-in later with neural patterns. The processes are therefore modelled on the human brain, but this framework is constructed through sets of declarative and procedural statements that might even be retrieved from an ontology or a database. How neural patterns form into concepts is not of interest, where the modelling is at the conceptual level to start with. The original model was a 3-level architecture [14][12] that included a bottom optimising level to construct patterns, through a middle aggregating level to an upper cognitive level that would perform some level of thinking. Later papers [7][6] then suggested a close relationship between a pattern ensemble and a hierarchical structure that may be modelled as a Concept Tree [5][11]. This 'unit of work' maps quite closely to the lower pattern-forming and the middle aggregating levels of the original cognitive model. These structures are also thought to be knowledge-based, or formed more through rote-learning. There was then originally a time-based layer that joined the bottom two levels with the upper cognitive level. In fact, this layer has been replaced by a types-based layer and the timing has been made implicit in the new upper-level network structure and concept tree instances. The cognitive level is thought to be more experience-based, making use of the learned knowledge in arbitrary ways, to build up a picture of the preferences for the intelligent agent. The upper level is also thought to be quite like the human brain cortex region [17]. For this paper, the cognitive layer has been re-designed, but it has different functions in the human brain as well and so it is not restricted to only one design. One version of this 3-level architecture has therefore been modelled now, but as a framework only. It is possible to see how the structure forms and how the different levels would interact, although there is still not enough information to show how real intelligence might be derived from it. Because the structure is built from statistical processes, it is assumed that intelligence will require flexibility that could be realised through layers of simple mechanisms applied over the complex network structure [29]. It is possible that the brain layers not only apply a different function, but also represent a different type of information, whereby the actual meaning of the wiring between neurons changes. As always, the construction process must be generic, it cannot be specific to any one problem or domain and so it must be relatively simple in nature. Then to enrich the knowledge and add context, different types of link, neuron and even a modular setup are considered, to allow the brain to reason over more complex information, but generated from the same sources.

The rest of the paper is organised as follows: section 2 describes the modular architecture that has been implemented, including mapping to the original 3-level architecture. Section 3 describes how the Cognitive Process Language [6] has been translated over to the neural architecture. Section 4 gives some implementation details on the computer program that has been written. Section 5 gives some related work and a discussion on the biological aspects of the model in particular, while section 6 gives some conclusions to the work. The Appendixes give some more examples of how the CPL can be represented in the network.

## 2    Modular Architecture

As stated, this paper describes implementation details for a distributed and neural framework. The framework does not contain all of the functionality required for intelligent thinking, but it places the concepts that it learns into appropriate context, so that other algorithms can extend the structure and more easily make use of the information. The concepts are learned at a coarse level of granularity. There is no process for neural clusters to form concept groups by themselves, but rather, the information and structure is presented to the network in parts and the network is able to autonomously build its own structure from each part. There are at least two different modules involved, where the higher-level cognitive processes require a different learning strategy to the lower level memory of objects and object relations. Following the design in [6] and [7], these object memories make use of neural ensembles and also hierarchies, represented by Concept Trees [5][11]. These are built by presenting to the network, parts of an already known ontology [15], where the network also knows the sub-concept relation for each element in the ontology. This is therefore only a statistical process of adding the ontology parts correctly, but the construction process adds the information in slightly different formats, making the knowledge-retrieval possibilities more interesting. The upper cognitive level is built differently to the Symbolic Neural Network design of [13]. In particular, the network is constructed from the Cognitive Process Language (CPL) of [6], which requires a different structure from clustering memory objects. Again, the CPL is already defined and it is mainly a matter of adding it to the network correctly. There is no real reasoning or ambiguity about how the network gets constructed. Construction of the cognitive level is described in section 3, while construction of the memory objects is described in the following sections.

### 2.1    Linking Flexibility

The problem of adding flexibility (or plasticity) to the structure is answered by using a geometrically progressive architecture that also makes use of the different element attributes, but at a very generic level. The concept trees and ensemble, for example, can be used to store information about static objects, where these objects have some type of natural order to their situation. For example, a car drives on a road and so it may be a sub-concept of road. Process rules probably require a different structure, because the rules may use a different object ordering based on concept 'use' and not concept situation. Therefore, different modules are appropriate, but the modules still have to represent the same concept types. Then, when the concept is triggered in one module, the same concept may need to be triggered in a related module. What would be interesting is if the trigger in a related module is for the concept type only and not the exact same instance. If considering the real brain, then the type might relate to frequency, signal strength, or some other general property, but that can be sent between modules. And now the concept is being represented and linked to in two ways – values for actual instances and values for generic types. If one module wants to know about names, for example and asks the next module to match with the name pattern, it might not receive the exact instance that the request originated from. This might be an advantage, where the first module would then run its process over different types of knowledge that would be related, but would produce different types of result. If the search can extend anywhere, then a success measurement is that it completes a cycle, when the original input criteria are still being met. As each input event also has a timestamp, then a third way to record the event is to store the structure for the timestamp itself and not try to integrate it into a value-based network. This was the suggestion for the bottom level of the symbolic neural network [13] and it is used again in this paper for the concept trees (see also section 3.2).

## 2.2    Geometric Flexibility

Along with linking flexibility, there can be a geometric progression through the 3-level architecture. The ensembles are placed in the bottom optimising level and would be deep nested structures that represent whole domains of knowledge. This is achieved by combining ensemble parts when they overlap or when one part is contained in another part. Note that ensembles can nest concepts, which is very similar to tree branches, but not exactly the same. Nested patterns occupy the same space but maybe do not have the directional links of a tree. With the ensemble, it can also store multiple instances of a pattern or concept type at any level and so it can store what is presented exactly. The Concept Trees have been related to the ensemble as part of an ensemble-hierarchy structure [7], where the same nodes in each, link directly with each other (binding) and may even oscillate when there is a match. This paper does not model that idea, but it still provides a direct link from the ensemble node 'type' to its related node in the concept tree. The concept tree therefore only stores one instance of each node type at a level and so when that aggregated or abstract representation is activated, it sends a signal to all nodes of that type in the ensemble, again to give that generic flexibility. Another difference is that the concept tree is created based on event instances or timestamps. Each input event gets stored as a separate concept tree, along with whatever structure is in the input and a link key is added to the structure to relate it with the underlying ensemble. All of the concept trees remain separate however unless there is containment. If one tree is contained completely in another tree, then the two can be merged and the link key sets updated. If there is only overlap between trees, then they are not merged. This leads to more shallow and smaller trees that represent input instances and at a more abstract level. The top cognitive level in the architecture actually stores a different type of information for this paper. It is a procedural description of how the use the basic concept sets. Preferences or learned rules can be described by Horn clauses [13], for example. These clauses are a further abstraction that represent each desired concept, simply as a concatenation that has been learned from somewhere. The Horn clause could also make better use of single node representations for what it wants to link with. This is actually quite sensible in a biological sense. The object memories are more likely to be static entities with multiple links between each and so they would be best suited to the deep trees. As discussed later in section 5, the Concept Trees begin to look a bit like the psychological phenomenon of chunking. Chunking is where the brain stores smaller pieces of information that are easier to digest and the concept trees can actually span more than 1 ensemble and so spread any search process over several domains of knowledge. The Horn clause would be used in the cortex region and would want to change concepts about more quickly. Therefore, a minimal representation that can be easily replaced would be preferable.

## 3    Translating the Cognitive Process Language

A Cognitive Process Language was suggested in [6] as a way of describing actions rather than objects in a script. The language is purposely simplistic and can only be used to create a useful framework at the moment, although, the level of detail has not been studied very seriously. The language is similar to RDF, where it describes each process step as a triple. The 3 elements of the triple describe: the object in question, some effector or trigger on that object and the source of the effector. Trying to map this over shows that the language translates quite nicely to a hierarchical structure, but that it is not able to maintain the natural ordering of the ontology objects. This is not surprising, because the objects are being used by another entity (a human, for example) in a different context each time. Therefore, the network probably requires a different construction method to add processes than to add static objects. A method has in fact been worked out that can add the process steps in a consistent and generic way and the resulting framework is interesting because objects of interest can be found by searching for shortest-path cycles that contain the objects. The algorithm for adding process events is as follows:

1.  Form a base list of all of the process step triples declared in the script.
2.  Form a bottom layer in the (symbolic) network that is 3 nodes only.
    a.  Each node is a concatenation of one of the entity types in each triple.
    b.  So, the first node is all of the objects, the second node is all of the effectors and the third node is all of the sources.
3.  Form a layer above this that creates new nodes, where each node shares concepts with more than 1 of the entity type nodes.
4.  Form a new layer above this that links each of the original triple nodes to the shared concept nodes of the second layer, that it shares a concept with.
5.  Form a new layer above this that creates new nodes for any more shared concepts in the triple layer. If a concept has already been used for a node, then it cannot be used again.
6.  If any of the triples do not link with a shared layer, then they can be linked to the types layer instead.

If some concepts are dead ends then they are not part of cycles and are only to introduce the required objects into the process. Other concepts can traverse several cycles and the most relevant ones for a concept can be found by searching for the shortest cycle routes in the network, where at least some of the nodes contain the concept.

## 3.1    Cook an Egg Example

The example in the paper [6] of cooking an egg has been translated over to the network structure, using the algorithm just described. Further examples can be found in Appendix A. It should be noted that the instruction set is not very detailed and a more complete study might try to determine how detailed it is allowed to be. The constructed network is therefore only a framework for representing the concepts and not a full set of instruction for how to cook an egg. The following steps were declared for the problem:

1. Go into the kitchen and get the pot.
2. Go to the tap in the kitchen and fill the pot with water.
3. Go to the cooker in the kitchen and switch the heat on.
4. Place the egg in the pot and the pot on the cooker heat and wait for it to boil.
5. Monitor the heat level.
6. Wait for the egg to cook.

Resulting in the following triples rules, with the order of object-effector-source:

| | | |
|---|---|---|
| 1.  P·D·K. | 4.  E·W·P. | 6.  E·H·P. |
| 2.  P·W·T. | 5.  P·H·B. | 7.  B·H·P. |
| 3.  H·B·C. | | |

The order is the same in every triple and the following symbols were used: Kitchen (K), Cupboard (D), Tap (T), Water (W), Cooker (C), Hob (B), Heat (H), Pot (P) and Egg (E).

The constructed network is shown in Figure 1. To construct this type of network means that an entity instance must play more than 1 role in the script. For example, it must be both an object and an effector or source, and so there must be some crossover that way. This is very reasonable, when thinking that the result of one process is often the input to another one. If thinking about Water, for example, the shortest cycle includes the nodes {W, EWP, P, PWT} which represent the concepts of Water, Egg, Pot and Tap. Egg requires a similar cycle of {E, EHP, P, EWP} for the concepts Egg, Heat, Water and Pot. The action to remove the pot from the cupboard in the kitchen (PDK) is added as a dead end and not as part of a cycle. To use it therefore could mean to trace to the Pot node once only, which is what the action is for. With this setup, the network does not really extend beyond 4 or 5 levels, which is interesting (see the biological section 5). The original triple instances are all used by the second linking level. Other examples of implementing this algorithm can be found in Appendix A.

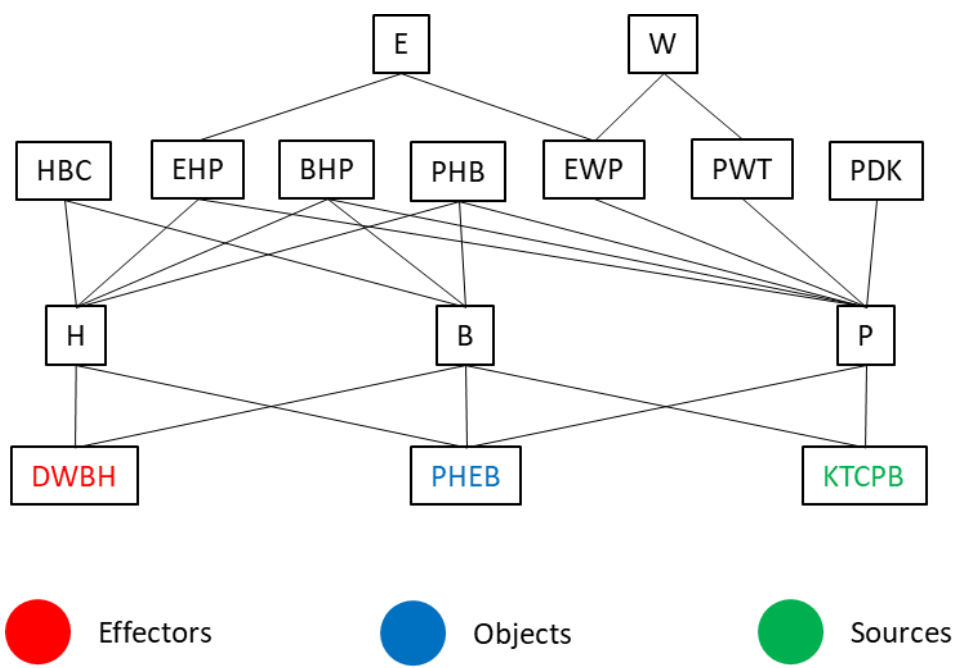

Figure 1. Symbolic Neural Network created from the CPL for Cook an Egg.

## 3.2    Scheduling the Tasks

Shortest paths can therefore help to define what concepts go together, but there is no clear order for how those concepts might be used. An earlier paper tried to tackle the problem of scheduling [10] by using a nested structure that would fire inwards. When a layer was activated, it would fire inhibitors outwards to switch its neighbouring layer off. The idea was that each layer would also link somewhere else and could therefore activate some other region in order. The original idea for the cognitive model top level was that there would be a hierarchy of concepts from simple to more complex ones and that the very top global concepts, representing real-world entities, would trigger each other, to start a process of thinking. The problem with this is that it was decided that the trigger links were basically the same as any other link and so the only thing of interest would be the hierarchy structure. That in itself is fine and is the established theory for how the neocortex arranges its information [25][17], but this paper is able to suggest an alternative format for storing more procedural knowledge. One hierarchy may be about what a recipe is, while another may be about how to cook it, for example. It was discovered however that individual instances can help to reduce noise and the concept trees make use of that idea in this paper.

While the network structure does not have an ordering, searching for paths may also help to schedule the tasks. The first thing would be to consider the leaf nodes as the starting points for the thinking. Then obviously cycling will encourage activation and so maybe a cycle starting from Water or Egg will find the Pot node. In fact, the Water node will also find the Tap node. A person thinking this way would realise that Pot and Tap are currently missing and so before the Water cycle can be completed, these nodes need to be realised. The Pot is found in the Cupboard and so the outer-most level of the scheduling network should try to activate Tap and the PDK node first. Once the Pot node is found, it can be used in the cycles concerning Egg and Water. There might be ambiguity with regard to Heating the Egg and Pot, or putting Water in the Pot, but in fact, the order is clear. Putting Water in the Pot is part of the Water and Tap cycle and so must occur first. Therefore, a clear task order can be realised from the search process and this order may be aligned more, back to the natural order of the environment again. Starting at the leaf nodes and trying to realise the concepts in order would therefore result in the following for the network in Figure 1.

1. Kitchen – Cupboard – Pot.
2. Tap - Water.
3. Cooker – Hob – Heat.
4. All concepts now realised and so any cycles can be traversed.

So, this would be another region on-top of the network that is learnt from the path rules and it would derive the order in which the rules should be applied, from entities that are missing as part of the cycles. If this new top layer then pairs with the network nodes (binding), it could prepare the nodes for later activation, which is known to be a feature of real brain networks. For this paper and probably this problem, a set of shallow linked trees, as in [6], figure 2 is preferred. Another option was mentioned in [10] that was nested regions that would fire in order from outer to inner, but if the ordering is always clear, then linked paths will suffice. An example scheduling layer is shown in Figure 2. It is probably OK to link to the base nodes that would be Tap or Hob, because they are the sources for the effector.

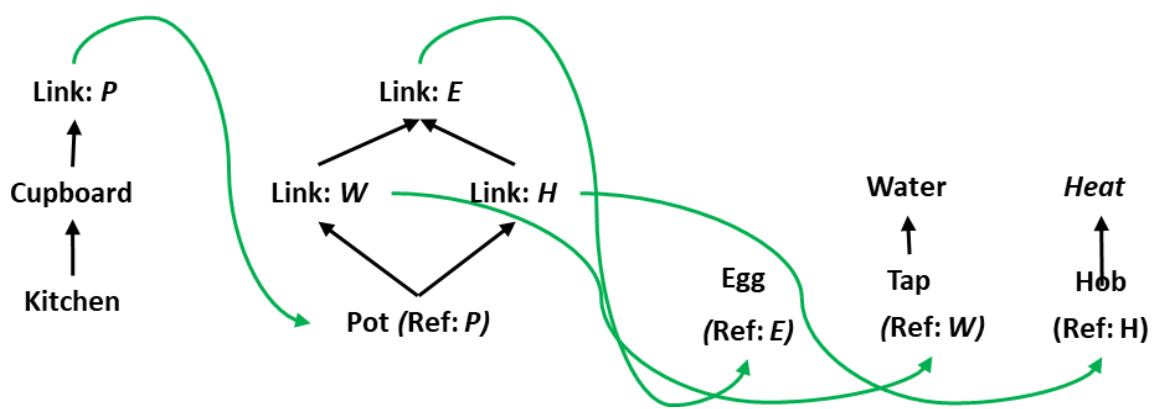

Figure 2. Top-Level Sparse Scheduler for the Cognitive Network.

There still remains the problem of how the two modules (bottom two levels and the upper level) would join through links and it may be the case that if types can be aggregated, there is a layer of separate instances for each concept type. Then the two modules can use this layer to link and transfer information. This leads to an overall architecture shown in the schematic of Figure 3.

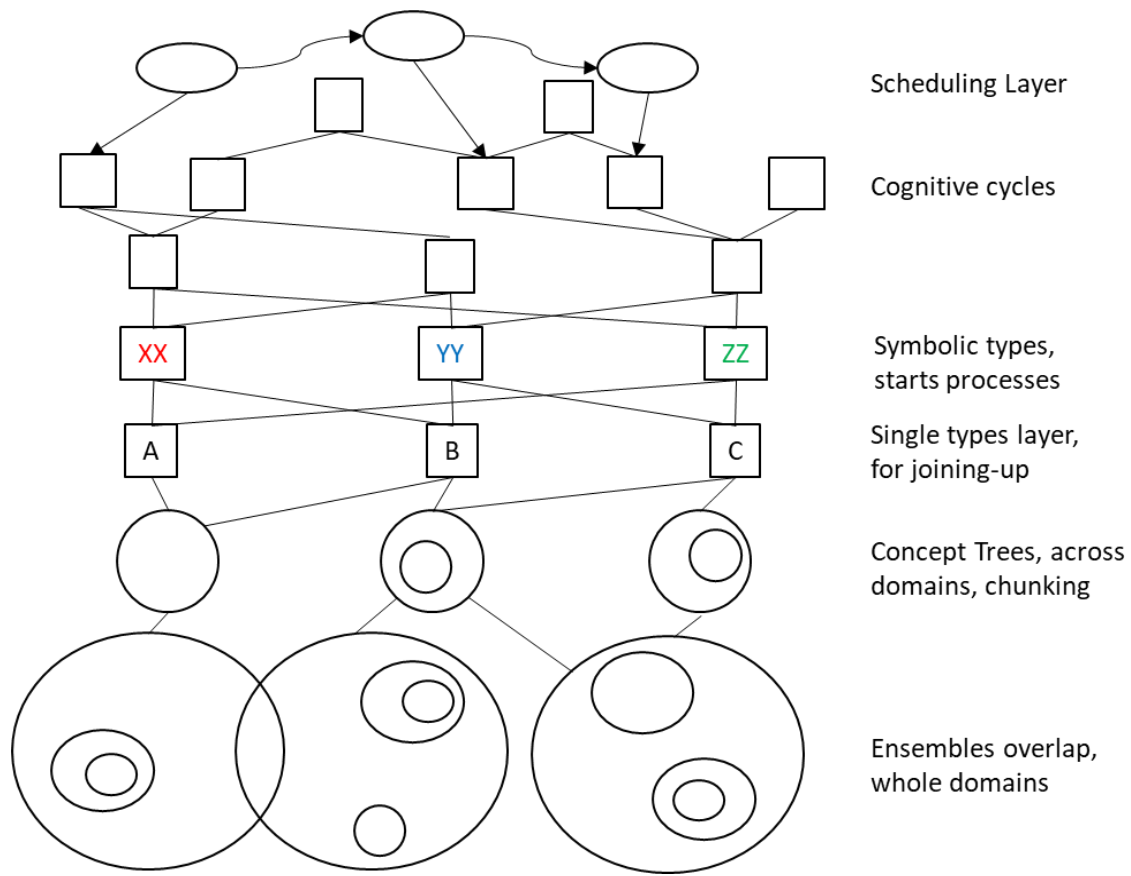

Figure 3. Schematic of the whole architecture.

## 4    Implementation Details

A computer program has been written in the Java language to read XML scripts that describe ontologies or the Cognitive Process Language as sets of nested elements. For the ontology, only part of it is read each time and is then presented to the network. The ontology was created from the example in [28] that is a smart home for sensor monitoring. The network then re-constructs the ontology, using the correct nested ordering. If the parts presented are random, then there can be variation in what structure is finally built. For the CPL, the whole language is presented to the Symbolic Neural Network and it is a matter of constructing the structure correctly from the algorithm described in section 3.

## 4.1   Ensemble View

This is the bottom-level view that is concepts nested in other concepts. The nesting should be done on pattern type only and each 'event' is added as a whole new ensemble. The nesting is really also a tree structure and so there is a definite structure to match with, where matching is done on the symbol value. If a new part is contained in an existing part, then new key sets can be added to the existing part instead of creating new nodes. If two parts overlap, then they can be combined and new keys or nodes added as appropriate. Using a house description as an ontology, where the different rooms could be different domains, a simple test produced the following ensemble construct from the ontology, shown in Figure 4.

## 4.2   Concept Tree View

This is the middle-level view that is again concepts nested in other concepts and each 'event' is again added as a whole new tree. With the concept tree however, multiple instances of a type, at a level, are aggregated into a single node. The structure is also not split on value, but kept as a whole, in relation to the link or event key. This allows for structures wholly contained in other structures to be combined, but not to combine overlapping structures. While the instances are experience-based, if there is structure, then that is also knowledge-based. What is good is that the same structure that creates the ensemble can create the concept tree. The ensemble nodes are then able to link directly with it. The same simple test produced the concept tree constructs shown in Figure 4. If each room is thought to be a separate domain, then some instances span across domains.

| Ensembles | Concept Trees | Concept Trees |
|---|---|---|
| Home<br>  Living Room<br>    motion_sensors<br>      M011<br>    Bathroom<br>      items<br>    items<br>      TV<br>  Kitchen | Home [Link_10]<br>  Kitchen [Link_10]<br>    items [Link_10]<br>      Pot [Link_10]<br><br>Home [Link_4, Link_6]<br>  Living Room [Link_4]<br>  Kitchen [Link_4, Link_6]<br>    items [Link_4] | Home [Link_8]<br>  Bedroom [Link_8]<br>    items [Link_8]<br>      Bed [Link_8]<br><br>Home [Link_7]<br>  Bedroom [Link_7]<br>    motion_sensors [Link_7]<br>  Living Room [Link_7]<br>    motion_sensors [Link_7] |

| items<br>Sink<br>Pot<br>motion_sensors<br>M017<br>Bedroom<br>motion_sensors<br>items<br>Bed<br>Dining Room<br><br>Dining Room<br>items | Home [Link_3]<br> Living Room [Link_3]<br>  items [Link_3]<br>  TV [Link_3]<br><br>Home [Link_1, Link_2]<br> Living Room [Link_1, Link_2]<br>  motion_sensors [Link_1, Link_2]<br>  Bathroom [Link_1]<br>   items [Link_1]<br><br>Dining Room [Link_5]<br> items [Link_5] | M011 [Link_7]<br><br>Home [Link_6, Link_8]<br> Kitchen [Link_6, Link_8]<br>  motion_sensors [Link_6, Link_8]<br>  M017 [Link_6]<br><br>Home [Link_9, Link_10]<br> Dining Room [Link_9]<br> Kitchen [Link_9, Link_10]<br>  items [Link_9, Link_10]<br>  Sink [Link_9, Link_10] |

Figure 4. Ensemble construction from the presentation of random ontology parts.

### 4.2.1 Counting Rule

The Concept tree is supposed to be constructed using a basic rule that a parent node must have an equal or greater occurrence count than one of its sub-nodes. That has not really been considered in this paper but probably remains legitimate. If, for example, the Kitchen was being linked to and not the House, then there must be other sources for the Kitchen and a real-world link between Kitchen and House would be weakened from that. So, it would probably make sense to re-structure the concept tree to make Kitchen a base concept, and also reflect that in the ensemble somehow, so that the whole structure remains as more solid units.

### 4.3 Symbolic Neural Network

This is constructed from the Cognitive Process Language. It uses the same concept set as the ontology, but a different information set and the natural ordering does not have to be maintained. It is constructed by presenting the whole script and is then deterministic through a specific algorithm, so this part would not be stochastic in any way, where that type of behaviour would transfer over to how it might be used after it was constructed.

## 5    Related Work and Discussion

There have been lots of attempts at building systems that copy the human brain. Most of the currently successful designs, Deep Mind [26] in particular, model first at the neural level and learn weight sets from that, to cluster into concepts through the hierarchies. Other designs are described in [22], where one option, used in SPAUN [2], is to transform an ensemble mass into a vector-style structure, with weighted sets of concepts or features, but what appears to be missing from models is contextual information. This is made clear from one of the original designs called SOAR [21]. That system adheres strictly to Newell and Simon's physical symbol system hypothesis [27], which states that symbolic processing is a necessary and sufficient condition for intelligent behaviour. SOAR exploits symbolic representations of knowledge (called chunks) and uses pattern matching to select relevant knowledge elements. Basically, where a production matches the contents of declarative (working) memory the rule fires and then the content from the declarative memory is retrieved. SOAR still suffers from problems of memory size and heterogeneity. There is also the problem that production rules are not general knowledge but are specific and so there is not a real understanding at the symbolic level. IBM's Watson [18] is also declarative, using NLP and relies on the cross-referencing of many heuristic results (hyperheuristics) to obtain intelligent results. Context is a key feature of the Watson system. It may be that the cross-referencing of different feature types from the same information sets in this paper's model can provide some level of context.

Hawkins et al. [16][17] have tried to model the neocortex exactly. As quoted in an earlier paper [8], the paper [1] describes a theory that is quite similar. They call the framework the Specialized Neural Regions for Global Efficiency (SNRGE) framework. The paper describes that 'the specializations associated with different brain areas represent computational trade-offs that are inherent in the neurobiological implementation of cognitive processes. That is, the trade-offs are a direct consequence of what computational processes can be easily implemented in the underlying biology.' The specializations of the paper correspond anatomically to the hippocampus (HC), the prefrontal cortex (PFC), and all of neocortex that is posterior to prefrontal cortex (posterior cortex, PC). Essentially, prefrontal cortex and the hippocampus appear to serve as memory areas that dynamically and interactively support the computation that is being performed by posterior brain areas. The PC stores overlapping distributed representations used to encode semantic and perceptual information. The HC stores sparse, pattern separated representations used to rapidly encode ('bind') entire patterns of information across cortex while minimizing interference. The FC stores isolated stripes (columns) of neurons capable of sustained firing (i.e., active maintenance or working memory). They argue against temporal synchrony and prefer to argue for coarse-coded distributed representations (CCDR) ([19] and others) instead. The hippocampus is certainly another region mentioned for this model to and while this paper deals with an upper refactored level, clustering higher-level objects is another option.

The symbolic network can use Horn clauses to list a set of concepts to search for. The list would be of types that would link through the abstract nodes sets to actual type instances in the ensembles. If this has to traverse the concept trees first, then that can add further flexibility and restrictions, when the tree might even span to a different domain. For one thing, the type can relate to frequency, signal strength, or some other general property of a real brain that can be sent between modules. The nesting, for example, can relate to a wiring length or frequency and so a general rule would be that a larger pattern has longer connections and so can accommodate smaller patterns with shorter connections. The paper [23] has mapped the real brain and was able to show that the functional architecture of the brain can be mapped to linear structures that link across modules. They note that: 'The Maximal Spanning Forest reveals both intra- and inter-hemispherical modules, and the presence of small modules alongside with larger sub-networks.' and conclude that 'In addition the chain-like feature of the basal modules facilitate the wiring and reconnection processes, not asking for a specific region as a hub, but functioning the whole module as a multiple access point for the functional plasticity.'

## 5.1   Chunking

The design is being modelled at the concept level, which makes it a declarative design and so it is probably not possible to map it to cellular layers known about in the neocortex [25][17]. But it can be noted that the neocortex layers contain a more-sparse layer at the top with horizontal links and then layers that repeat function below. A comparison with chunking [24] may be more appropriate. In cognitive psychology[1], chunking is a process by which individual pieces of an information set are broken down and then grouped together in a meaningful whole. The chunks by which the information is grouped is meant to improve short-term retention of the material, thus by-passing the limited capacity of working memory. A chunk is a collection of basic familiar units that have been grouped together and stored in a person's memory. These chunks are able to be retrieved more easily due to their coherent familiarity. It is believed that individuals create higher order cognitive representations of the items within the chunk. The items are more easily remembered as a group than as the individual items themselves. These chunks can be highly subjective because they rely on an individual's perceptions and past experiences, that are able to be linked to the information set. The size of the chunks generally range anywhere from two to six items, but often differ based on language and culture. According to Johnson [20], there are four main concepts associated with the memory process of chunking: chunk, memory code, decode, and recode. The chunk is a sequence of to-be-remembered information that can be composed of adjacent terms. These item or information sets are to be stored in the same memory code. The process of recoding is where one learns the code for a chunk, and decoding is when the code is translated into the information that it represents. Chunking is the concept trees. That is coded through translating to single item types and then to type clusters. That is then decoded back into the specific type instances of the CPL. If re-coding is when the scheduling layer makes use of the new network structure, then it maps across almost exactly. It would also suggest that the coded parts can be used by more than one specific process.

The idea of chunking is varied and uses different models for different brain region or functionality. While the cortex might start off as short-term memory, there is also a model for long-term memory. In that case, a chunk can be defined as 'a collection of elements having strong associations with one another, but weak associations with elements within other chunks' [4]. But again, the associations are key and if they are missing, an expert may perform as poorly as a beginner, which was demonstrated by a memory test between master chess players and beginners. It is the cross-referencing that also produces the flexibility and this how the brain remembers what is significant, when cycles complete. The cross-referencing in further levels of the symbolic network requires that a concept is used in more than 1 context. For the CPL mapping, a concept would have to be in two or more of 'object, effector or source'. This might produce a contrast or difference for that concept that the brain can again recognise and use as a disruptive source from where to learn. The psychology of Gestalt was used in an earlier paper [7] with relation to the ensemble-hierarchy model and it is supposed to describe how the sum of the parts is not only greater than the whole but may be different to it. This has been challenged in a recent paper [3] that describes: 'In cognitive psychology, grouping has been mainly studied by two traditions: (a) Gestalt psychology, which focused on perception, although other aspects of cognition such as problem solving were considered as well, and (b) the line of research interested in chunking, which focuses on memory, although aspects of perception and problem solving have been studied as well.' This is not an attempt to define chunking, but another interesting point may be when it is used for creating long-term memory: 'The likelihood of remembering information is increased by (a) making associations, either with other items to remember or with prior knowledge, (b) elaborating information by processing it at a deep level, for example by making semantic associations, (c) recoding, where information is represented in a different and hopefully more efficient way, (d) creating retrieval structures (LTM items that are organised in such a way that encoding and retrieving new information are optimised), and (e) grouping information (i.e. chunking). Note that mnemonics nearly always use LTM.' The paper concludes that Gestalt may be correct when realised as the sum of the parts 'plus their relations', when it looks more like chunking. So, the two theories have similarities. If Gestalt theory is concerned more with perception and processes, then it would fit more into the CPL network and that network structure is somehow re-organised and holistic, rather than the separate but linked fragments in the lower memory structures.

---

[1] See Wikipedia, https://en.wikipedia.org/wiki/Chunking_(psychology).

## 6   Conclusions

This paper has described implementation details for a new cognitive model that has been developed over a number of papers. The system has studied the problem at the neural level earlier, but the current work can describe a whole system at the declarative level only. The bottom two levels of the cognitive model are essentially the same as in the original design. One small difference changes the modality of the concept trees, to represent type-based instances (with structure) instead of knowledge-based values. But this can complement the ensembles that store the knowledge and give flexibility over what might get searched. For this paper, the cognitive layer has been re-designed. An earlier design proposed the traditional hierarchical clustering of concepts and included a time-based layer between the modules. This paper proposes a new refactored network, with a types-based layer in-between, where the timing is made implicit in the network. But the human cortex is modular and so it is not restricted to only one design. While networks of arbitrary complexity can be learned, they are built from fully-defined scripts to start with, but once the information has been transferred to the network, it looks quite a lot like a brain model that can organise and reason over information by itself. If the script does not look quite right, then the network structure can even be a guide to what should be changed and so it may even be able to help with that construction process. This is still only a starting point from which to study this problem and it opens up many different possibilities. Comparisons with the biological world are important, because the human brain is still the model of intelligence that we are trying to copy.

## Appendix A: Examples of the CPL Converted to a Network

Two examples of converting a CPL into the network structure are shown here. A basic algorithm is described first, with the CPL rules constructed from that, followed by the network structure. As with the main paper example, blue is the objects, red is the effectors and green is the sources.

### 1. Drive a Car

Driver get the car
U + GC -> UCG

Start engine to drive
D + CE -> DEC

Keys to start the engine
K + EI -> KIE

Move the car with the pedal
M + CP -> MPC

Foot on pedal to move the car
M + PF -> MFP

Steer the car with the wheel
R + CW -> RWC

Hands on wheel to steer
H + RW -> HWR

Adjust speed
S + PF -> SFP

Driver (U), Garage (G), Car (C), Keys (K), Ignition (I), Engine (E), Foot (F), Pedal (P), Steering Wheel (W), Hands (H), Drive(D), Steer (R), Speed (S).

The upper scheduling layer could look like:

1. User - Garage - Car.

2. Keys – Ignition - Engine.

3. Foot – Pedal.

4. Hands – Steering Wheel.

5. All concepts now realised and so any cycles can be traversed.

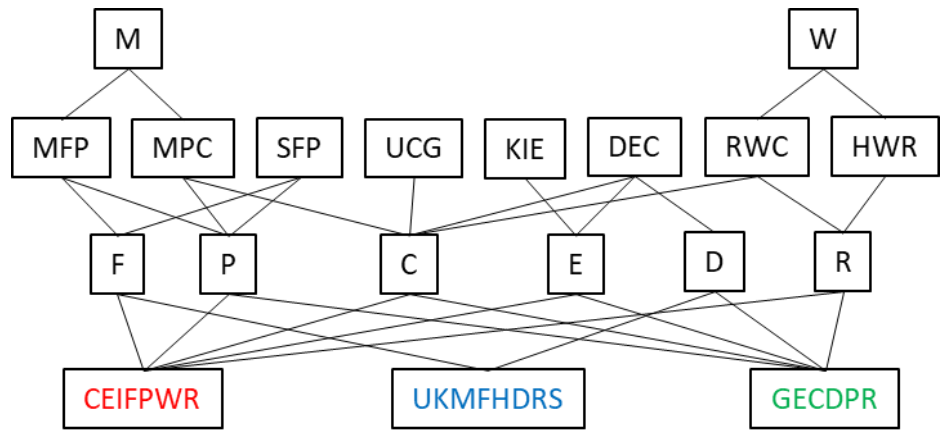

**2. Book a Holiday**

Log onto Internet from Computer
U + CI -> UIC

Find best price
P + HD -> PDH

Find Holiday web sites
W + IS -> WSI

Enter details to Book
B + HE -> BEH

Find Destination on Web Site
H + WD -> HDW

User (U), Computer (C), Internet (I), Search (S), Web Site (W), Holiday Web Site (H), Destination (D), Best Price (P), Booking Details (B), Enter Details (E).

The upper scheduling layer could look like:

1. User – Computer - Internet.

2. Internet - Search.

3. Web Site – Holiday Web Site.

4. Determine Price.

5. All concepts now realised and so any cycles can be traversed.

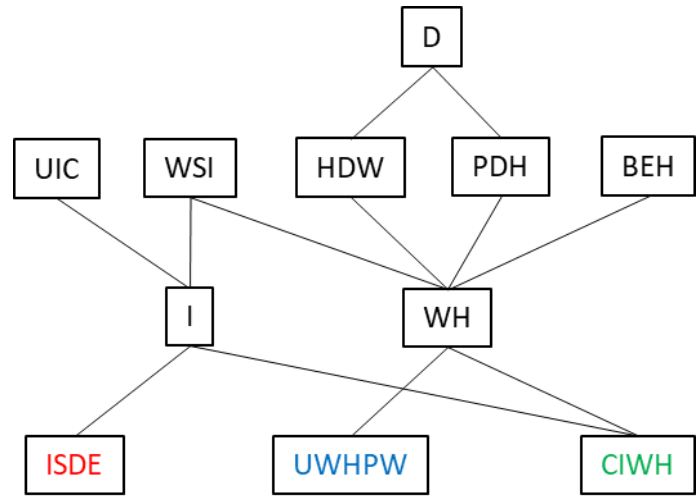